\documentclass[letterpaper, 10 pt, conference]{ieeeconf}  % Comment this line out if you need a4paper

\IEEEoverridecommandlockouts                              % This command is only needed if 
\usepackage{graphics} % for pdf, bitmapped graphics files
\usepackage{amsmath} % assumes amsmath package installed
\usepackage{amssymb}  % assumes amsmath package installed
\usepackage{graphicx}
\usepackage{subfiles}
\usepackage{multirow}
\usepackage{makecell}
\usepackage{tabularx}
\usepackage{soul}
\usepackage{color}
\usepackage[final]{changes}
\usepackage{caption}
\usepackage{hyperref}
\usepackage{cleveref}
\usepackage[T1]{fontenc}

\crefformat{section}{\S#2#1#3} % see manual of cleveref, section 8.2.1
\crefformat{subsection}{\S#2#1#3}
\crefformat{subsubsection}{\S#2#1#3}

\title{\LARGE \bf
Exploiting Task Tolerances in Mimicry-based Telemanipulation
}

\author{Yeping Wang$^{1}$, Carter Sifferman$^{1}$, and Michael Gleicher$^{1}$% <-this % stops a space
\thanks{$^{1}$Yeping Wang, Carter Sifferman, and Michael Gleicher are with the Department of Computer Sciences, University of Wisconsin-Madison, Madison 53706, USA
$\quad \quad \quad \quad \quad \quad \quad \quad \quad \quad \quad \quad \quad \quad \quad \quad $ 
{\tt\small [yeping|sifferman|gleicher]@cs.wisc.edu}}%
\thanks{This work was supported by Los Alamos National Laboratory and \added{the} Department of Energy, a University of Wisconsin Vilas Associates Award, and National Science Foundation award 1830242.}% <-this % stops a space
}

\makeatletter
\let\@oldmaketitle\@maketitle% Store \@maketitle
\renewcommand{\@maketitle}{\@oldmaketitle% Update \@maketitle to insert...
   \vspace{5mm}
    \includegraphics[width=7in]{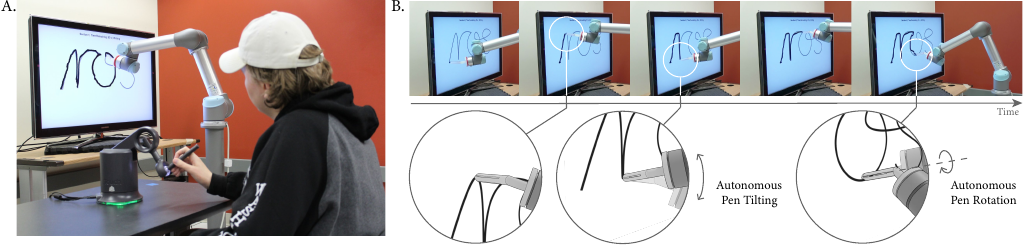}
    \captionof{figure}{We explore how autonomous robot adjustments within task tolerances shape task performance and user experience in mimicry-based telemanipulation. For example, (A) in a teleoperated writing task, our human-subject experiment results indicate that (B) if the robot autonomously tilts or rotates the pen within task tolerances to improve pen tip accuracy and motion smoothness,  telemanipulation is enhanced by the high-quality robot motions enabled by task tolerances, despite users lacking full control of the robot.}
    \label{fig: teaser}
    }
\makeatother

\begin{document}

 \vspace{-8mm}
\maketitle

\addtocounter{figure}{-1} % hack, otherwise following figures start with Fig 3

%%%%%%%%%%%%%%%%%%%%%%%%%%%%%%%%%%%%%%%%%%%%%%%%%%%%%%%%%%%%%%%%%%%%%%%%%%%%%%%%
\begin{abstract}
We explore task tolerances, \textit{i.e.}, allowable position or rotation inaccuracy, as an important resource to facilitate smooth and effective telemanipulation. Task tolerances provide a robot flexibility to generate smooth and feasible motions; however, in teleoperation, this flexibility may make the user's control less direct. In this work, we implemented a telemanipulation system that allows a robot to autonomously adjust its configuration within task tolerances. We conducted a user study comparing a telemanipulation paradigm that exploits task tolerances (\emph{functional} mimicry) to a paradigm that requires the robot to exactly mimic its human operator (\emph{exact} mimicry), and assess how the choice in paradigm shapes user experience and task performance. Our results show that autonomous adjustments within task tolerances can lead to performance improvements without sacrificing perceived control of the robot. Additionally, we find that users perceive the robot to be more under control, predictable, fluent, and trustworthy in functional mimicry than in exact mimicry. 
\end{abstract}

\section{Introduction}
Mimicry-based telemanipulation maps a human operator’s hand movement to a robot’s end effector in real time \cite{rakita2017motion, rolley2018bi}. 
The robot is often required to mimic the operator's movement as exactly as possible, so we call this paradigm \textit{exact} mimicry. \replaced{However, this approach may cause the robot to lose manipulability or generate jerky motions b}{B}ecause of the kinematic and dynamic differences between the robot and the operator\deleted{, exact mimicry may cause the robot to lose manipulability or generate jerky motions}. Exact mimicry may be overly \textit{restrictive} for some tasks, imposing unnecessary requirements on the robot.

Many \deleted{everyday} tasks are designed to be accomplished while allowing position or rotation inaccuracy, \textit{i.e.}, tolerances.
For instance, a writing task requires accurate pen tip positions, but does not require the pen to be strictly perpendicular to the surface\replaced{. Similarly, the rotation of a welding torch around its axis does not negatively impact the quality of a welding task \cite{de2017cartesian}. T}{, so wrist movements in writing can naturally rotate or tilt the pen.
Similarly, t}he aperture of a mailbox is generally wider than an envelope, avoiding the need for accurate alignment to drop a letter. 
These task-specific tolerances are naturally utilized by humans. 
In principle, these tolerances also give a robot extra freedom that it can exploit to better execute tasks, \textit{e.g.}, while writing, a robot can autonomously tilt the pen to avoid joint limits or singular configurations. However, in teleoperation, such autonomous adjustments \replaced{mean that the user lacks}{make the user lack} full control of the robot. The less direct control could harm user experience and task performance.

In this paper, we explore task tolerances as an important resource to facilitate \textit{functional} mimicry in telemanipulation. 
The functional mimicry paradigm allows a robot to autonomously adjust within tolerances to generate more accurate, smooth, and feasible motions. 
In our previous work, we presented \textit{RangedIK} \cite{wang2023rangedik} as a real-time motion generation method that exploits flexibility afforded by task tolerances. In this paper, we apply \textit{RangedIK} in a mimicry-based telemanipulation system and investigate whether task performance and user experience will be improved given autonomous robot adjustments within task tolerances. 

We conducted a human-subject experiment in which participants perform tasks using a teleoperation system with or without task tolerance exploitation. Our results indicate that autonomous robot adjustments within task tolerances lead to equal or better performance without sacrificing perceived control of the robot. Moreover, the participants perceived the robot was more under control, predictable, fluent, and trustworthy when it exploited task tolerances than when it exactly followed the users' commands. 

The central contribution of this paper is empirical evidence showing that exploiting the flexibility in task tolerances enables a robot to generate more accurate, smooth, and feasible motions, leading to better task performance and user experience in mimicry-based telemanipulation.

\section{Related Works}
We review relevant prior research from three areas: shared control methods, user perception of shared control systems, and semi-constrained tasks.

\subsection{Shared Control} \label{sec:shared_control}
Niemeyer et al. \cite{niemeyer2016telerobotics} organize teleoperation control methods in a spectrum that involves direct and shared control. Direct control allows the user to unambiguously control the robot, while shared control seeks to provide \added{some} autonomous assistance, such as obstacle avoidance \cite{you2012assisted, kang2021rcik, takayama2011assisted, acharya2018inference, milliken2017modeling} or guidance to an effective motion or strategy \cite{dragan2013policy, nikolaidis2017human}.
The teleoperation paradigm explored in this paper aims to gain benefits from both direct and shared control: it makes autonomous robot adjustments for more accurate, smooth, and feasible motions while preserving directness in the users' control. Below, we further articulate the distinctions between our method and prior shared control works.

Both our system and some shared control works autonomously control a subset of an end-effector's degrees of freedom (DoFs). However, our system exploits the DoFs that are \textit{task-irrelevant} to improve robot motion quality, while shared control systems often take over \textit{task-relevant} DoFs to reduce a user's workload, \textit{e.g.}, autonomously rotating a manipulator's gripper for grasping \cite{stoyanov2018assisted, abi2016visual}. 
While most shared control systems \textit{explicitly} assist task completion, the teleoperation paradigm explored in this paper provides \textit{implicit} assistance by generating high-quality robot motions.

\subsection{User Perception in Shared Control}
\label{sec:perception_shared_control}
While improving safety and performance, shared control may make the user's control less direct, suggesting a trade-off between performance and perception \cite{javdani2015shared}. While some prior works have shown that user satisfaction of a shared control system strongly correlates with task performance \cite{dragan2013policy, hauser2013recognition}, You et al. \cite{you2012assisted} found that users are willing to tolerate loss of control and less predictable motions \textit{only} for significant performance improvements. Moreover, Nikolaidis et al. \cite{nikolaidis2017human} found trust to be inconsistent with performance in some shared control systems.
In our teleoperation paradigm, although users lose some control of the robot by allowing autonomous adjustments within task tolerances, we anticipate that users will still feel in control because the adjustments do not affect task completion and are natural to users because humans also utilize task tolerances in manipulation.

To further understand the perception of shared control systems, prior research has studied intrinsic user qualities, \textit{e.g.}, Locus of Control (LoC). LoC describes the degree to which people believe that their behaviors affect the outcome of events in their life \cite{rotter1966generalized}. People with a more \textit{internal} LoC believe they have more control over the events in their life whereas people with a more \textit{external} LoC believe the outcome of events is more affected by external forces such as luck or fate. 
Prior works have found that people with a high internal LoC have trouble giving up control to an autonomous system\replaced{ \cite{takayama2011assisted, acharya2018inference}.}{. Takayama et al. \cite{takayama2011assisted} found that a collision avoidance system significantly increased task completion time for operators with a high internal LoC. Similarly, Acharya et al. \cite{acharya2018inference} found that people with high internal LoC were more likely to issue conflicting commands that diverge from an obstacle avoidance algorithm.} \replaced{Therefore, we anticipate that users with a high internal LoC are more sensitive to the autonomous robot adjustments in our \textit{functional} mimicry paradigm, leading to less improvements in task performance and user perception.}{In our study, we also anticipate users with a high internal LoC to have trouble giving up control to our autonomous system.}

\subsection{Utilizing Flexibility in Semi-Constrained Tasks}

A task with tolerances is semi-constrained, as it does not impose \added{strict} constraints to all six degrees of freedom of the end effector\added{, constructing a null space in which the robot can freely move}. Prior works have exploited the flexibility in semi-constrained tasks by projecting a secondary task into the null space of a semi-constrained task. Common secondary tasks include self-collision avoidance \cite{petrivc2011smooth}, singularity avoidance \cite{nemec2000null}, compliant behavior \cite{sadeghian2013task}, balancing a humanoid robot \cite{henze2016passivity, abi2018humanoid}, or conveying emotions \cite{claret2017exploiting}. In contrast to previous work, our robot exploits the flexibility in a semi-constrained task to generate accurate and smooth motions and we explore its effects in mimicry-based telemanipulation.

To generate motions that utilize flexibility in semi-constrained tasks, several semi-constrained motion planners have been presented \cite{Descartes, de2017cartesian, malhan2022generation, berenson2011task, cefalo2020opportunistic}. 
While semi-constrained motion planners can effectively generate motions for path following, they are not appropriate in time-sensitive scenarios such as teleoperation. In this work, we employ \textit{RangedIK}\cite{wang2023rangedik}, which is a per-instant pose optimization method, to exploit task tolerances and generate robot motions in real time.

\section{Functional Mimicry}
In this section, we introduce \textit{functional} mimicry in contrast to the commonly-used \textit{exact} mimicry, discuss task tolerances that are exploited by functional mimicry, and describe how \textit{RangedIK} enables functional mimicry. 

\subsection{Mimicry-based Telemanipulation}
\label{sec:mimicry}
A mimicry-based telemanipulation system maps the six degree-of-freedom (DoF) movement of a user’s hand to the robot’s end effector in real time. Such mimicry-control methods have been shown to be intuitive and effective in many applications \cite{rakita2017motion, rolley2018bi}. In this paper, we call these methods \emph{exact} mimicry because the robot \emph{tries} to match \textit{all} six DoFs. In reality, 6-DoF matching is often challenging because of the kinematic and dynamic differences between a robot and its human operator. Moreover, some DoFs are actually task-irrelevant and do not require high accuracy. 

While exact mimicry treats both task-relevant and task-irrelevant DoFs uniformly,  \textit{functional} mimicry deliberately sacrifices the accuracy in task-irrelevant DoFs to improve the accuracy in task-relevant DoFs. Specifically, functional mimicry allows a robot to autonomously move in task-irrelevant DoFs to generate smooth and feasible motions. 

\subsection{Task Tolerance}
Throughout \replaced{this}{the} paper, task tolerances are specified as the \deleted{allowable }amount of position or rotation inaccuracy \added{allowed} to complete a manipulation task. Task tolerances can facilitate task completion, enabling humans or robots to conduct the task without accurately manipulating all six DoFs. We describe task tolerances using the allowable inaccuracy in each DoF (examples are in Table \ref{tab:task_ranges}).

A task with tolerances often involves objects that have rotational symmetry or large effective regions. Rotational symmetry occurs when an object is equivalent under any rotation about a certain axis. Common objects with rotational symmetry include pens, bottles, and bowls. Rotational symmetry simplifies not only the manufacturing process (\textit{e.g.}, using a lathe) but also the usage of an object, \textit{i.e.}, users can manipulate one fewer rotational DoF. The rotational DoF about the rotational symmetry axis has unbounded tolerances. 

Aside from rotational symmetry, large effective areas also create task tolerances. 
The effective parts, or working areas, of many daily objects are designed to be larger than needed, such as the edge of a knife or the aperture of a letter box. Even a pen has a large spherical effective area on the pen tip to ensure enough contact with the writing surface when the pen is tilted. The position or rotational DoFs in the effective areas have bounded tolerances. 

\subsection{Exploiting Task Tolerance}
\label{sec:exploiting}

In this section, we describe how we use \textit{RangedIK} \cite{wang2023rangedik} to exploit task tolerances to enable functional mimicry. To perform a task with tolerances, the DoFs of a robot's end-effector can be classified into three categories: (1) a DoF with zero tolerance requires the robot to accurately match a \textit{specific} goal, \textit{e.g.}, a writing task requires accurate pen tip positions; (2) a DoF with unbounded tolerance provides the robot a \textit{range} of equally valid goals, such as allowing a pen to rotate about its principal axis; and (3) a DoF with bounded tolerance gives the robot a \emph{range} of acceptable goals with a preference toward a \emph{specific} goal, \textit{e.g.}, allowing a tilting pen but preferring it to follow user's commands. 
\textit{RangedIK} is a  real-time optimization-based motion synthesis method that is able to accommodate these three categories of requirements within a single, unified framework. 
With other objectives to avoid self-collision, maintain manipulability, keep joint positions within limits, and minimize joint velocities, accelerations, and jerks, \textit{RangedIK} enables a robot to exploit task tolerances to generate accurate, smooth, and feasible motions. 

\section{User Study}
Our evaluation was focused on assessing how automatic robot adjustments within task tolerances may influence robot telemanipulation. 
We pre-registered\footnote{\url{https://osf.io/2ryng/?view\_only=a04dac33e23c4448883f79302de0d0b1}} our hypothesis, study design, measures, sample size, and statistical analyses before collecting data.
Our central hypothesis is that allowing a robot to exploit task tolerances in mimicry-based telemanipulation will lead to better task performance and user experience than requiring the robot to exactly mimic its operator.

\subsection{Experimental Design \& Conditions} \label{sec:desgin}

Our user evaluation followed a within-participants design, with each participant working with the robots in both conditions. We counterbalanced the order in which the two conditions were presented.

\emph{Exact Mimicry} (control condition): as described in \cref{sec:mimicry}, the robot tried to exactly mimic all six degrees-of-freedom of the participant's hand movements. 
% mimicry control \cite{rakita2017motion}

\emph{Functional Mimicry} (experimental condition): as described in \cref{sec:mimicry}, the robot mimicked the participant's hand movements but was allowed to autonomously adjust within task tolerances. 

\subsection{Experimental Tasks} \label{sec:tasks}
\begin{figure*} [!tb]
  \centering
  \includegraphics[width=7in]{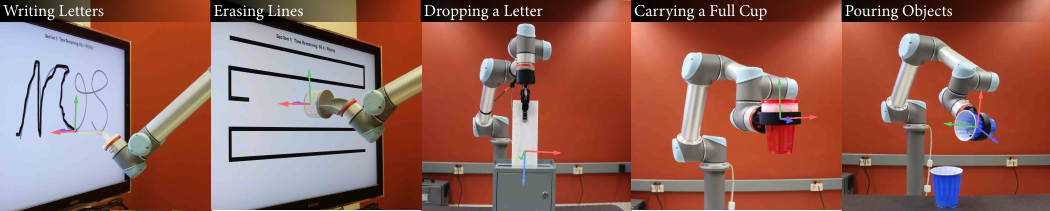}
  \caption{Our human-subject experiment evaluated \emph{exact} and \emph{functional} mimicry on five manipulation tasks that have tolerances. Table \ref{tab:task_ranges} lists the tolerances in the visualized coordinate frame, where red, green, and blue arrows represent $x$, $y$, and $z$ axes, respectively. 
    }
  \label{fig: tasks}
  \vspace{-5mm}
\end{figure*}

\begin{table}[!tb]
\caption{Task Tolerances}
\label{tab:task_ranges}
\vspace{-3mm}
\begin{center}
\begin{tabular}{lllllll}
\hline
\rule{0pt}{1.1\normalbaselineskip}%
Task & \multicolumn{6}{c}{Tolerances} \\
& \makecell[l]{$x$ \\ (m)} & \makecell[l]{$y$ \\ (m)} & \makecell[l]{$z$ \\ (m)} & \makecell[l]{rx \\ (rad)} & \makecell[l]{ry \\ (rad)} & \makecell[l]{rz \\ (rad)} \\[0.8mm] 
\hline
\rule{0pt}{1.1\normalbaselineskip}Writing Letters & 0 & 0 & 0 & $\pm\pi/6$ & $\pm\pi/6$ & $\pm\infty$ \\
Erasing Lines & 0 & 0 & 0 & 0 & 0 & $\pm\infty$  \\
Dropping Envelope & $\pm0.05$ & 0 & 0 & 0 & 0 & 0  \\
Carrying Full Cup & 0 & 0 & 0 & 0 & $\pm\infty$ & 0  \\
Pouring Objects & 0 & 0 & 0 & 0 & $\pm\infty$ & 0  \\[0.8mm] 
\hline
\vspace{-8mm}
\end{tabular}
\end{center}
\end{table}

We constructed five manipulation tasks that have tolerances (Figure \ref{fig: tasks} and Table \ref{tab:task_ranges}). The first two tasks involved manipulation on a whiteboard and a haptic stylus provided haptic feedback to assist participants in keeping the end-effector on the whiteboard plane. The other three tasks involved large robot movements in free space and we used a VR Controller as the input device. 

1) \emph{Writing Letters}: This task required participants to write letters on a virtual whiteboard. To ensure fair comparisons, we instructed participants to trace the same pattern as quickly as possible while maintaining accuracy. This task required accurate pen tip positions but \replaced{allowed}{allows} the pen to tilt (rotational tolerances about the $x$ and $y$ axes) or freely rotate about the pen’s principal $z$ axis. This task was challenging because it required the pen tip to accurately trace a long curve.

2) \emph{Erasing Lines}: With a round eraser as the robot's end-effector, participants wiped down a pre-defined, lawnmower trace on the whiteboard. Participants were instructed to finish the task as quickly as possible. 
The round eraser had rotational symmetry and the robot was free to rotate about the axis of the eraser.
This task required dexterously rotating the eraser to keep it parallel to the whiteboard while moving the eraser along a long, continuous path.

3) \emph{Dropping an Envelope}: This task required dropping an envelope into a mailbox. Participants were instructed to finish the task as quickly as possible but avoid collisions with the mailbox which bends the envelope. The aperture of the mailbox is 0.22 m long, enabling positional tolerance along the $x$ axis.  

4) \emph{Carrying a Full Cup}: Like holding a full cup of water, participants were instructed to keep a cup upright and carry the cup to a target position as quickly as possible. A warning sound was played if the cup was tilted more than 10 degrees. Keeping the cup upright required dexterous manipulation of the rotation about the $x$ and $z$ axes, while the rotational symmetry of the cup allowed unbounded rotational tolerance about the $y$ axis.

5) \emph{Pouring Objects}: This task involved pouring 10 binder clips from one cup to another cup on the tabletop. Participants were instructed to pour all the binder clips as quickly as possible. The pouring task required smooth rotation to prevent spilling. 

\subsection{Implementation Details}

A participant's hand motions were captured using a 3D Systems Touch Stylus at approximately 100 Hz for the \textit{Writing} and \textit{Erasing} tasks and an HTC Vive Controller at approximately 60 Hz for the \textit{Dropping}, \textit{Carrying}, and \textit{Pouring} tasks. The hand motions were mapped to the end-effector of a six degree-of-freedom Universal Robots UR5 manipulator. 
We used the open-source implementation of \textit{RangedIK}\footnote{\url{https://github.com/uwgraphics/relaxed\_ik\_core/tree/ranged-ik}} to generate robot motions \added{at 200 Hz} in both conditions. As described in \cref{sec:exploiting}, the autonomous robot adjustments in functional mimicry were enabled by handling the degrees of freedom (DoFs) with tolerances differently. Meanwhile, the exact mimicry condition required the robot to accurately match all DoFs of the end-effector.
A button on the stylus and the trigger button on the Vive controller served as clutching buttons to \replaced{connect or disconnect}{build or demolish connections} to the robot. To efficiently display curves for participants to trace or erase, we implemented a virtual whiteboard using the \texttt{pygame} Python library on a TV with a resolution of 1920 $\times$ 1080 and a screen size of 1.05m $\times$ 0.59m. 

\subsection{Experimental Procedure}
Upon receiving consent, an experimenter introduced the goal of the study and the usage of both input devices (a haptic stylus and a VR controller) to participants. Then participants were presented the first condition, in which participants performed practice tasks and then the experimental tasks described in \cref{sec:tasks}. The practice tasks were simplified from the experimental tasks: writing a straight line, erasing a small area, dropping an envelope into a mailbox that is close to the robot, carrying a full cup for a short time, and pouring binder clips into a cup that is close to the robot. 
Upon finishing all the practice and experimental tasks, participants filled out a questionnaire regarding their experience in the condition. 
This procedure, including performing the practice and experimental tasks and filling out the questionnaire, was repeated for the other condition. Upon finishing both conditions, participants completed a demographic questionnaire followed by an  Internal Control Index questionnaire \cite{duttweiler1984internal} to measure their Locus of Control. 
The experiment ended with a semi-structured interview.
% in which participants were asked three questions: \emph{``Which section do you prefer?'', ``Can you tell the differences between sections?'', ``Throughout the experiment, have you ever felt that the robot is no longer under control?''}.
Participants received \$10 compensation for about 40 minutes in the experiment.

\subsection{Measures}
We employed a combination of objective and subjective measures to assess participants’ performance and user experience.
\subsubsection{Objective measures}
We employed completion time and an error metric to assess the performance of each task. The maximum time limit to complete each task was 60 seconds. Because the time and the error metric were possibly associated, \textit{e.g.}, a participant who hurriedly finished the task in a short time might have a large error, we formulated a \textit{combined metric} for each task to aggregate the data.  We combined task time $T$ and error metrics $E$ by normalizing them\replaced{ over all participants}{ to $[0,1]$} and summing them together. 
\begin{equation}
    \textit{Combined Metric} = \frac{T-T_{min}}{T_{max}-T_{min}} + \frac{E-E_{min}}{E_{max}-E_{min}}
\end{equation}
The resulting range is $[0,2]$, where a lower value indicates better performance. 

For the \textit{Writing} task, we formulated a trajectory error metric to assess how well participants trace the target curve. The trajectory error metric is the sum of an accuracy score and a completeness score. The accuracy score is the average error distance between the pen tip and its closest point on the target curve. The closest point on the target curve is marked as reached. To compute the completeness score, we first associate an arc-length parameter value in $[0,1]$ to all points on the target curve. The completeness score is the maximum arc-length parameter value of the points that are marked as reached. We normalize the accuracy score \replaced{over all participants. The resulting}{to $[0,1]$ and the} range of trajectory error is $[0,2]$, where a lower value indicates a better trajectory.

To measure the performance of the \emph{Erasing} task, we measured the area of non-erased marks and reported them in $m^2$. For the \emph{Dropping} task, we counted the number of collisions between the envelope and the drop box, which lead to bends in the envelope. We counted the amount of time when the cup was tilted more than 10 degrees to measure errors in the \emph{Carrying} task. For the \emph{Pouring} task, we counted the number of clips that fell outside of the target cup or stayed in the robot's cup.

In addition, we measured the \emph{accuracy}, \emph{smoothness}, and \emph{manipulability} of robot motions using 6 metrics. Motion \emph{accuracy} was measured using mean position error (m) and mean rotation error (rad) between an end-effector's pose and its goal pose specified by the user. The errors were measured only in the task-relevant degrees of freedom. We used mean joint velocity (rad/s), mean joint acceleration (rad/s$^2$), and mean joint jerk (rad/s$^3$) to assess motion \emph{smoothness}. Motion \emph{manipulability} was measured by mean Yoshikawa manipulability \cite{yoshikawa1985manipulability}, where a higher value indicates better manipulability.

\subsubsection{Subjective measures}
We administered a questionnaire based on prior research in mimicry-based telemanipulation \cite{rakita2020effects} and shared control \cite{javdani2015shared, qiao2021learning} to measure perceived \textit{control}, \textit{predictability},  \textit{fluency}, and \textit{trust}. Additionally, we employed NASA TLX \cite{hart1988development} to assess perceived workload.

\begin{table}[tb]
% \begin{tabularx}{\textwidth}{X}
\caption{Items in Subjective Questionnaire}
\label{tab:questions}
\vspace{-3mm}
\begin{center}
\begin{tabular}{l}
\hline
\rule{0pt}{1.1\normalbaselineskip}%
\textbf{Control} \\
\hspace{5mm} I felt in control. \\
\hspace{5mm} I felt I could control the robot. \\
%\hline
\textbf{Predictability} \\
\hspace{5mm} The robot consistently moved in a way that I expected. \\
\hspace{5mm} The robot’s motion was not surprising. \\
\hspace{5mm} The robot responded to my motion inputs in a predictable way. \\
\hspace{5mm} I was often confused about where to move the robot. \\
%\hline
\textbf{Fluency} \\
\hspace{5mm} The robot contributed to the fluency of the interaction. \\
\hspace{5mm} The robot and I worked fluently together as a team. \\
%\hline 
\textbf{Trust} \\
\hspace{5mm} I trusted the robot to do the right thing at the right time. \\
\hspace{5mm} The robot was trustworthy. \\
\hline
\end{tabular}
\end{center}
\vspace{-4mm}
\end{table}
    % \end{tabularx}
% \fi

\subsection{Participants} \label{sec:participants}

We recruited 20 participants from a university campus (10 females, 9 males, and 1 non-binary). Participants, aged 18 to 39 (M = 23.40, SD = 5.03), had a variety of education backgrounds, including engineering, business, biology, and communication arts. Through 5-point Likert scale, participants reported low-to-moderate familiarity with robots (M = 2.35, SD = 1.06), 3D video games (M = 2.65, SD = 1.59), Computer-Aided Design (CAD) software (M = 2.15, SD = 1.06), and VR controllers (M = 2.40, SD = 1.20). One of 20 participants reported themselves as left-handed, with others being right-handed. Participants used their dominant hands to operate the robot.

Participants' Locus of Control (LoC) was measured using the Internal Control Index (ICI) \cite{duttweiler1984internal}, which involves twenty-eight 5-point Likert scales and generates an ICI score ranging from 28 to 140. Following prior work \cite{chiou2021trust}, we equally divided ICI scores into three categories: \replaced{high \textit{external} LoC ($<65$),  \textit{average} LoC ($65-102$), and high \textit{internal} LoC ($>102$)}{scores lower than 65 have high \textit{external} LoC, scores between 66 and 102 have \textit{average} LoC, and scores larger than 102 have high \textit{internal} LoC}. In our study, 10 participants had average LoC and the remaining 10 participants had high internal LoC.

\begin{figure*} [tb]
  \centering
  \includegraphics[width=\textwidth]{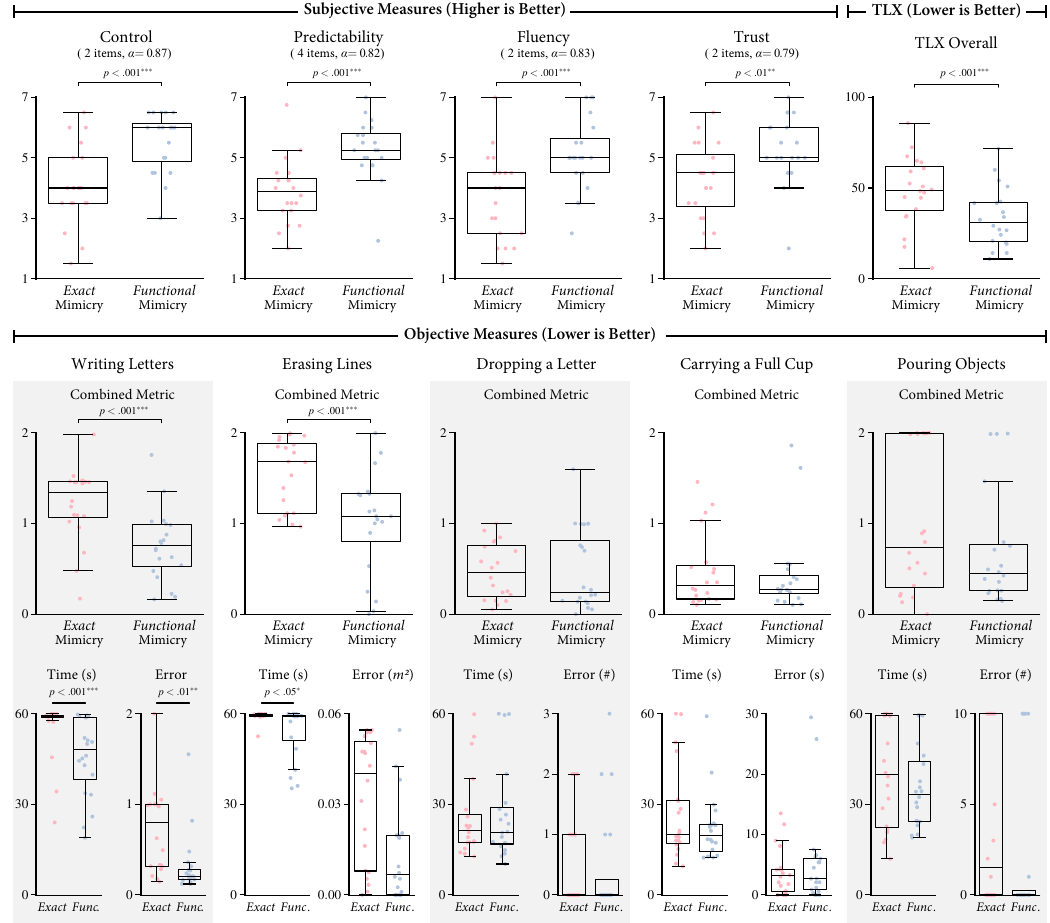}
  \caption{Box and whisker plots of data from the user perception and performance measures for our experiment. 
  The top and bottom of each box represent the first and third quartiles,
and the line inside each box is the statistical median of the data. The length of the box is defined as the interquartile range
(IQR). The whiskers are within a maximum of 1.5 IQR.
    TLX Overall means averaged NASA Task Load Index scores.
    }
  \label{fig: main_results}
%\vspace{-5mm}
\end{figure*}

\begin{table*}[tb]
\caption{Motion Qualities$^{\dag}$}
\label{tab:motion_results}
\vspace{-3mm}
\begin{center}
\begin{tabular}{llllllll}
\hline
\rule{0pt}{1.1\normalbaselineskip}%
Task & Method & \makecell{Mean Pos. \\ Error (m)} & \makecell{Mean Rot. \\ Error (rad)} & \makecell{Mean Joint \\ Vel. (rad/s)} & \makecell{Mean Joint \\ Acc. (rad/s$^2$)} & \makecell{Mean Joint \\Jerk (rad/s$^3$)} & \makecell{Mean Mani-\\pulability} \\
\hline

\rule{0pt}{1.1\normalbaselineskip}%
%%%%%%%%%%%%%%%%%%%% paste data from python %%%%%%%%%%%%%%%%%%%% 
\multirow{2}{*}{Writing Letters}& \emph{Exact} Mimicry & 0.091$\pm$0.085 & N/A$^{\ddag}$ & 0.133$\pm$0.05 & 1.80$\pm$0.7 & 53.9$\pm$19.5 & 0.067$\pm$0.02 \\   
& \emph{Functional} Mimicry & \textbf{0.006}$\pm$0.009 & N/A$^{\ddag}$ & \textbf{0.076}$\pm$0.04 & \textbf{0.45}$\pm$0.2 & \textbf{10.5}$\pm$5.1 & \textbf{0.085}$\pm$0.02 \\   
\hline 
\rule{0pt}{1.1\normalbaselineskip}%
\multirow{2}{*}{Erasing Lines} & \emph{Exact} Mimicry & 0.093$\pm$0.074 & 0.0204$\pm$0.007 & 0.240$\pm$0.11 & 3.11$\pm$1.8 & 91.1$\pm$56.3 & 0.060$\pm$0.02 \\   
& \emph{Functional} Mimicry & \textbf{0.025}$\pm$0.020 & \textbf{0.0107}$\pm$0.007 & \textbf{0.227}$\pm$0.08 & \textbf{2.23}$\pm$0.9 & \textbf{62.9}$\pm$26.1 & \textbf{0.081}$\pm$0.02 \\   
\hline 
\rule{0pt}{1.1\normalbaselineskip}%
\multirow{2}{*}{Dropping an Envelope}  & \emph{Exact} Mimicry & 0.090$\pm$0.148 & 0.0073$\pm$0.007 & 0.150$\pm$0.05 & 1.42$\pm$0.5 & 40.0$\pm$15.0 & 0.053$\pm$0.03 \\   
& \emph{Functional} Mimicry & \textbf{0.028}$\pm$0.042 & \textbf{0.0070}$\pm$0.004 & \textbf{0.149}$\pm$0.06 & \textbf{1.18}$\pm$0.5 & \textbf{32.3}$\pm$15.7 & \textbf{0.069}$\pm$0.03 \\  
\hline 
\rule{0pt}{1.1\normalbaselineskip}%
\multirow{2}{*}{Carrying a Full Cup}  & \emph{Exact} Mimicry & 0.053$\pm$0.093 & 0.0059$\pm$0.009 & 0.179$\pm$0.09 & 1.63$\pm$0.8 & 45.8$\pm$19.6 & 0.048$\pm$0.02 \\   
& \emph{Functional} Mimicry & \textbf{0.007}$\pm$0.004 & \textbf{0.0021}$\pm$0.001 & \textbf{0.116}$\pm$0.05 & \textbf{0.90}$\pm$0.4 & \textbf{24.3}$\pm$9.3 & \textbf{0.071}$\pm$0.02 \\   
\hline 
\rule{0pt}{1.1\normalbaselineskip}%
\multirow{2}{*}{Pouring Objects} & \emph{Exact} Mimicry & 0.133$\pm$0.144 & 0.0149$\pm$0.008 & 0.218$\pm$0.10 & 2.42$\pm$1.9 & 71.1$\pm$71.5 & 0.035$\pm$0.02 \\   
& \emph{Functional} Mimicry & \textbf{0.064}$\pm$0.065 & \textbf{0.0102}$\pm$0.005 & \textbf{0.158}$\pm$0.06 & \textbf{1.35}$\pm$0.7 & \textbf{35.9}$\pm$20.8 & \textbf{0.068}$\pm$0.02 \\    
\hline 
%%%%%%%%%%%%%%%%%%%%%%%%%%%%%%%%%%%%%%%%%%%%%%%%%%%%%%%%%%%% 
\multicolumn{8}{l}{\rule{0pt}{1\normalbaselineskip}%
$^{\dag}$ The range values are standard deviations. The better value between the two telemanipulation paradigms for each measure is highlighted in bold.} \\
\multicolumn{8}{p{0.95\linewidth}}{\rule{0pt}{1\normalbaselineskip}% 
$^{\ddag}$ The position and rotation errors were measured in the task-relevant degrees of freedom that do not have tolerances. In the writing task, all three rotational degrees of freedom had tolerances, so no rotation errors were measured.} 
\vspace{-5mm}
\end{tabular}
\end{center}
\end{table*}

\begin{table}[tb]
\caption{Statistical results of our measures}
\label{tab:main_results}
\vspace{-3mm}
\begin{center}
\begin{tabular}{l@{\hspace{7pt}}lll@{\hspace{10pt}}l@{\hspace{10pt}}l}
\hline
\rule{0pt}{1.4\normalbaselineskip}%
Metrics & \makecell{\textit{Exact}\\ Mimicry}  & \makecell{\textit{Functional}\\ Mimicry} &  \multicolumn{3}{l}{Statistical Test Results}  \\ 
 & Mean {\scriptsize(SD)} & Mean {\scriptsize(SD)} & $t(19)$ & \makecell[c]{$p$} & \makecell[c]{$d$} \\ 
\hline
\rule{0pt}{1\normalbaselineskip}%
%%%%%%%%%%%%%%%%%%%%%%%%%  Paste here %%%%%%%%%%%%%%%%%%%%%%%%%%%%
Control&4.08 {\scriptsize (1.27)}&\textbf{5.52} {\scriptsize (0.97)}& -4.68&$<.001$&1.29\\
Predictability&3.88 {\scriptsize (1.04)}&\textbf{5.29} {\scriptsize (0.96)}& -4.84&$<.001$&1.41\\
Fluency&3.75 {\scriptsize (1.38)}&\textbf{5.10} {\scriptsize (1.18)}& -3.97&$<.001$&1.05\\
Trust&4.22 {\scriptsize (1.21)}&\textbf{5.25} {\scriptsize (1.09)}& -3.47&$.003$&0.89\\
TLX Overall &48.3 {\scriptsize (18.9)}&\textbf{33.6} {\scriptsize (16.0)}& 4.10&$<.001$&0.84\\
Writing Metric&1.21 {\scriptsize (0.40)}&\textbf{0.75} {\scriptsize (0.38)}& 4.55& $<.001$&1.16 \\
Writing Error&0.72 {\scriptsize (0.46)}&\textbf{0.32} {\scriptsize (0.32)}& 3.29&$.004$&1.01 \\
Erasing Error&0.03 {\scriptsize (0.02)}&\textbf{0.01} {\scriptsize (0.02)}& 3.82&$.001$&1.01 \\
Dropping Metric&\textbf{0.47} {\scriptsize (0.30)}&0.48 {\scriptsize (0.44)}& -0.06&.952&0.02 \\
Pouring Metric&0.99 {\scriptsize (0.77)}&\textbf{0.69} {\scriptsize (0.62)}& 1.24&.230&0.43 \\
Pouring Time&40.1 {\scriptsize (17.1)}&\textbf{35.3} {\scriptsize (13.2)}& 0.88&.391&0.31 \\
\hline
\rule{0pt}{1\normalbaselineskip}%
& Mdn {\scriptsize(IQR)} & Mdn {\scriptsize(IQR)} & W & \makecell[c]{$p$} & \makecell[c]{$r$} \\ 
\hline
\rule{0pt}{1\normalbaselineskip}%
Writing Time&59.2 {\scriptsize (0.88)}&\textbf{48.0}  {\scriptsize (0.10)}& 176&$<.001$&0.84\\
Erasing Metric&1.68 {\scriptsize (0.77)}&\textbf{1.08}  {\scriptsize (0.21)}& 174&$<.001$&0.83\\
Erasing Time&59.5 {\scriptsize (0.71)}&\textbf{59.2}  {\scriptsize (0.4)}& 116&.020&0.61\\
Dropping Time&21.5 {\scriptsize (9.10)}&\textbf{20.8}  {\scriptsize (11.6)}& 28&.622&0.13\\
Dropping Error& 0.00  {\scriptsize (1.00)}& 0.00  {\scriptsize (0.25)}& 4&.809&0.09\\
Carrying Metric&0.32 {\scriptsize (0.37)}&\textbf{0.27}  {\scriptsize (0.26)}& 16&.784&0.08\\
Carrying Time&19.9 {\scriptsize (14.1)}&\textbf{19.8}  {\scriptsize (6.18)}& 64&.245&0.30\\
Carrying Error&3.15 {\scriptsize (3.68)}&\textbf{2.70}  {\scriptsize (5.48)}& 41&.332&0.27\\
Pouring Error&1.50 {\scriptsize (10.0)}&\textbf{0.00}  {\scriptsize (0.25)}& 30&.164&0.45\\
%%%%%%%%%%%%%%%%%%%%%%%%%%%%%%%%%%%%%%%%%%%%%%%%%%%%%%%%%%%%%%%%%%%
\hline

\multicolumn{6}{p{3.15in}}{\rule{0pt}{1\normalbaselineskip}For normally distributed data, we report the mean value with standard deviations, $t$-test results, and Cohen's $d$. For non-normally distributed data, we report the median with interquartile ranges, Wilcoxon signed rank test results (W is the signed-rank sum), and rank biserial correlation coefficient $r$ \cite{king2018statistical}.}
\vspace{-8mm}
\end{tabular}
\end{center}
\end{table}

\subsection{Results}
We first determined whether our data had normal distributions using the Shapiro–Wilk normality test. For each measure, if the result of the Shapiro-Wilk test suggested that the differences between conditions were normally distributed ($p>.05$), we employed the two-tailed paired $t$-test to evaluate the difference between conditions. 
If the \textit{p}-value of the Shapiro-Wilk test was smaller than $.05$, we could not assume normality and used the two-tailed Wilcoxon signed rank test. 
Additionally, we calculated Cohen's $d$ or matched-pairs rank biserial correlation coefficient \cite{king2018statistical} to assess the effect size of normally or non-normally distributed data, respectively. 
Figure \ref{fig: main_results} and Table \ref{tab:main_results} summarize our results. 

\emph{Task Performance}
Our results indicate that in the \textit{Writing} and \textit{Erasing} tasks, the participants performed significantly better (both $p<.001$) in functional mimicry than in exact mimicry with large effect sizes ($d=1.16$ for \textit{Writing} and $r=.83$ for \textit{Erasing}). The effect sizes of the other three tasks were not large enough to lead to statistically significant differences with the sample size in our study. Our results show that the autonomous adjustments in task tolerances do not lead to the detriment of task performance and can improve the performance of some tasks.

\emph{User Perception}
Our results revealed that participants still perceive the robot to be under control even though the robot did not exactly mimic their movements.  
In particular, the robot that exploits task tolerances was perceived to be significantly more under control, predictable, fluent, and trustworthy, and to require significantly lower workload (all $p$-values $<.01$) with large effect sizes (all Cohen's $d$s $>.8$). 

In the post-experiment interview, 18 out of 20 participants stated that they preferred the functional mimicry robot. Participants described functional mimicry in different ways, for example,
P3 commented that ``\textit{I thought that [the functional mimicry robot] compensated more for my movements, which at first I didn't like, but then once I seemed to get a better feel for the compensation, it went more smoothly.}'' 
We speculate that the compensation refers to the autonomous robot adjustments within task tolerances.
Moreover, the subtle adjustments could increase perceived fluency, as described by P11: ``\textit{I felt that [the functional mimicry robot] was more forgiving when I made a mistake. Like it was able to recover and continue with the task without making big adjustments.}'' 

\emph{Motion Qualities} As shown in Table \ref{tab:motion_results}, when functionally mimicking users, the robot generated more accurate (fewer position and rotation errors) and smoother (lower velocities, accelerations, and jerks in the joint space) motions with better manipulability compared to the robot that exactly mimicked users, across all the five tasks. The numerical motion accuracy and smoothness matched the perceived predictability and fluency reported by the participants. Although prior work \cite{wang2023rangedik} has already shown that task tolerances can be exploited to generate high-quality motions, in this experiment, the robot was controlled by a user in real time, in contrast to following a pre-defined trajectory in prior work. Our results indicate that, in mimicry-based telemanipulation, autonomous robot adjustments within task tolerances enabled higher-quality motions, leading to user experience and performance improvements.

\section{Discussion}
In this work, we investigated functional mimicry as a smooth and effective telemanipulation paradigm. Functional mimicry allows a robot to exploit flexibility in task tolerances to generate more accurate, smooth, and feasible motions. Our user evaluation demonstrated that the autonomous adjustments within task tolerances led to equal or better performance without sacrificing perceived control of the robot. Moreover, the functional mimicry manipulator that exploited task tolerances was perceived to be more under control, predictable, fluent, and trustworthy than the manipulator that exactly mimicked its human operator. Below, we discuss additional findings, limitations of this work, and implications for future teleoperation systems.

\subsection{Subgroup Analysis}

\begin{figure} [tb]
  \centering
  \includegraphics[width=3.4in]{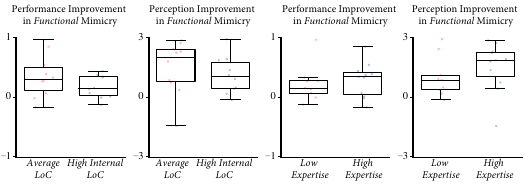}
  \vspace{-5mm}
  \caption{Our subgroup analyses explore the effects of functional mimicry on different types of users. The performance improvements are the differences between the combined objective metrics in each paradigm, which are averaged over the five tasks. The perception improvements are the differences between subjective measures in each condition, which are averaged over the four subjective scales. For both the performance and perception improvements, a more positive value indicates a larger improvement in the functional mimicry condition compared to the exact mimicry condition.}
  \label{fig: subgroup_results}
  \vspace{-5mm}
\end{figure}

Figure \ref{fig: subgroup_results} shows our subgroup analysis results. As mentioned in \cref{sec:perception_shared_control}, prior works have found that people with a high internal Locus of Control (LoC) have \replaced{trouble}{problems} giving up control to an autonomous system. We observed a similar phenomenon in our study. Although not statistically significant, the autonomous adjustments provided in functional mimicry led to less performance and perception improvement in high internal LoC participants than in participants with an average LoC. Moreover, we also classified participants' expertise according to their reported familiarity in operating robots, avatars in video games, and virtual objects in Computer-Aided Design software or virtual reality. We identified 9 participants with low expertise and 11 with high expertise. Although not statistically significant,  functional mimicry brought a larger performance and perception improvement in expert users than in non-expert users. We note that the result is not directly comparable with what is found in prior shared control work \cite{milliken2017modeling}, where non-expert users benefit more from autonomous assistance. As described in \cref{sec:shared_control}, in prior work the shared control system \textit{explicitly}  assists users in avoiding obstacles and directly contributes to task completion, so non-expert users \replaced{with insufficient}{of low} obstacle avoidance skills gain more benefits from the explicit assistance. Meanwhile, our system \textit{implicitly} assists them by generating high-quality robot motions. While both expert and non-expert users benefit from the high-quality motions, we speculate that expert users can take more advantage of the high-quality motions to complete the manipulation tasks. 

\subsection{Limitations}
While our results demonstrate the potential for functional mimicry to provide smooth and effective telemanipulation, the limitations of the present work suggest directions for future research.
In our empirical study, participants and the robot were located in the same room, which gives the participants perfect situational awareness of the robot and its surroundings and allows the robot to be controlled with low latency. With limited situational awareness (\textit{e.g.}, viewing the robot's workspace through a camera) and high latency, human operators may be more sensitive and disturbed by autonomous robot adjustments, even if they are within task tolerances. Future work should evaluate functional mimicry when the human operator and the robot are in separate physical spaces. While this work demonstrates the benefits of exploiting task tolerances, the task tolerances in our experiment were manually specified. Future work should investigate algorithms to automatically detect task tolerances. 

\subsection{Implications}
Instead of forcing a robot manipulator to exactly mimic its human operator, a mimicry-based teleoperation system can allow some task-specific inaccuracy. Such flexibility in task tolerance can be exploited by a robot to generate accurate, smooth, and feasible motions, leading to user perception and performance improvements. We believe that \replaced{our interaction paradigm is beneficial to teleoperation of welding, sanding, painting, wiping, pouring, and many other tasks that allow some positional or rotational inaccuracy.}{many high-precision teleoperation scenarios would benefit from our interaction paradigm.} Additionally, our user study results suggest that autonomous robot adjustments do not impede perceived control of the robot as long as the end-effector poses are within task tolerances. This finding provides a robot more freedom to enhance motion generation. Aside from the improved accuracy, smoothness, and manipulability demonstrated in this work, we believe that other aspects of robot motions, such as legibility, predictability \cite{dragan2013legibility}, and expressiveness \cite{venture2019robot}, can be enhanced with the flexibility in task tolerances.

\vspace{-0.5mm}
\section{Acknowledgement}
We would like to thank Seth Peterson for 3D printing tool mounts for the robot used in the study\added{ and anonymous reviewers for their suggestions and comments}.

\bibliography{reference}

\begin{thebibliography}{10}
\providecommand{\url}[1]{#1}
\csname url@samestyle\endcsname
\providecommand{\newblock}{\relax}
\providecommand{\bibinfo}[2]{#2}
\providecommand{\BIBentrySTDinterwordspacing}{\spaceskip=0pt\relax}
\providecommand{\BIBentryALTinterwordstretchfactor}{4}
\providecommand{\BIBentryALTinterwordspacing}{\spaceskip=\fontdimen2\font plus
\BIBentryALTinterwordstretchfactor\fontdimen3\font minus
  \fontdimen4\font\relax}
\providecommand{\BIBforeignlanguage}[2]{{%
\expandafter\ifx\csname l@#1\endcsname\relax
\typeout{** WARNING: IEEEtran.bst: No hyphenation pattern has been}%
\typeout{** loaded for the language `#1'. Using the pattern for}%
\typeout{** the default language instead.}%
\else
\language=\csname l@#1\endcsname
\fi
#2}}
\providecommand{\BIBdecl}{\relax}
\BIBdecl

\bibitem{rakita2017motion}
D.~Rakita, B.~Mutlu, and M.~Gleicher, ``A motion retargeting method for
  effective mimicry-based teleoperation of robot arms,'' in \emph{Proceedings
  of the 2017 ACM/IEEE International Conference on Human-Robot Interaction},
  2017, pp. 361--370.

\bibitem{rolley2018bi}
E.-J. Rolley-Parnell, D.~Kanoulas, A.~Laurenzi, B.~Delhaisse, L.~Rozo, D.~G.
  Caldwell, and N.~G. Tsagarakis, ``Bi-manual articulated robot teleoperation
  using an external rgb-d range sensor,'' in \emph{2018 15th International
  Conference on Control, Automation, Robotics and Vision (ICARCV)}.\hskip 1em
  plus 0.5em minus 0.4em\relax IEEE, 2018, pp. 298--304.

\bibitem{de2017cartesian}
J.~De~Maeyer, B.~Moyaers, and E.~Demeester, ``Cartesian path planning for arc
  welding robots: Evaluation of the descartes algorithm,'' in \emph{2017 22nd
  IEEE International conference on emerging technologies and factory automation
  (ETFA)}.\hskip 1em plus 0.5em minus 0.4em\relax IEEE, 2017, pp. 1--8.

\bibitem{wang2023rangedik}
Y.~Wang, P.~Praveena, D.~Rakita, and M.~Gleicher, ``Rangedik: An
  optimization-based robot motion generation method for ranged-goal tasks,'' in
  \emph{2023 IEEE International Conference on Robotics and Automation
  (ICRA)}.\hskip 1em plus 0.5em minus 0.4em\relax IEEE, 2023.

\bibitem{niemeyer2016telerobotics}
G.~Niemeyer, C.~Preusche, S.~Stramigioli, and D.~Lee, ``Telerobotics,''
  \emph{Springer handbook of robotics}, pp. 1085--1108, 2016.

\bibitem{you2012assisted}
E.~You and K.~Hauser, ``Assisted teleoperation strategies for aggressively
  controlling a robot arm with 2d input,'' in \emph{Robotics: science and
  systems}, vol.~7.\hskip 1em plus 0.5em minus 0.4em\relax MIT Press USA, 2012,
  p. 354.

\bibitem{kang2021rcik}
M.~Kang, Y.~Cho, and S.-E. Yoon, ``Rcik: Real-time collision-free inverse
  kinematics using a collision-cost prediction network,'' \emph{IEEE Robotics
  and Automation Letters}, vol.~7, no.~1, pp. 610--617, 2021.

\bibitem{takayama2011assisted}
L.~Takayama, E.~Marder-Eppstein, H.~Harris, and J.~M. Beer, ``Assisted driving
  of a mobile remote presence system: System design and controlled user
  evaluation,'' in \emph{2011 IEEE international conference on robotics and
  automation}.\hskip 1em plus 0.5em minus 0.4em\relax IEEE, 2011, pp.
  1883--1889.

\bibitem{acharya2018inference}
U.~Acharya, S.~Kunde, L.~Hall, B.~A. Duncan, and J.~M. Bradley, ``Inference of
  user qualities in shared control,'' in \emph{2018 IEEE Int. Conf. on Robotics
  and Automation (ICRA)}.\hskip 1em plus 0.5em minus 0.4em\relax IEEE, 2018,
  pp. 588--595.

\bibitem{milliken2017modeling}
L.~Milliken and G.~A. Hollinger, ``Modeling user expertise for choosing levels
  of shared autonomy,'' in \emph{2017 IEEE International Conference on Robotics
  and Automation (ICRA)}.\hskip 1em plus 0.5em minus 0.4em\relax IEEE, 2017,
  pp. 2285--2291.

\bibitem{dragan2013policy}
A.~D. Dragan and S.~S. Srinivasa, ``A policy-blending formalism for shared
  control,'' \emph{The International Journal of Robotics Research}, vol.~32,
  no.~7, pp. 790--805, 2013.

\bibitem{nikolaidis2017human}
S.~Nikolaidis, Y.~X. Zhu, D.~Hsu, and S.~Srinivasa, ``Human-robot mutual
  adaptation in shared autonomy,'' in \emph{2017 12th ACM/IEEE Int. Conf. on
  Human-Robot Interaction (HRI)}.\hskip 1em plus 0.5em minus 0.4em\relax IEEE,
  2017, pp. 294--302.

\bibitem{stoyanov2018assisted}
T.~Stoyanov, R.~Krug, A.~Kiselev, D.~Sun, and A.~Loutfi, ``Assisted
  telemanipulation: A stack-of-tasks approach to remote manipulator control,''
  in \emph{2018 IEEE/RSJ International Conference on Intelligent Robots and
  Systems (IROS)}.\hskip 1em plus 0.5em minus 0.4em\relax IEEE, 2018, pp. 1--9.

\bibitem{abi2016visual}
F.~Abi-Farraj, N.~Pedemonte, and P.~R. Giordano, ``A visual-based shared
  control architecture for remote telemanipulation,'' in \emph{2016 IEEE/RSJ
  International Conference on Intelligent Robots and Systems (IROS)}.\hskip 1em
  plus 0.5em minus 0.4em\relax IEEE, 2016, pp. 4266--4273.

\bibitem{javdani2015shared}
S.~Javdani, S.~S. Srinivasa, and J.~A. Bagnell, ``Shared autonomy via hindsight
  optimization,'' \emph{Robotics science and systems: online proceedings}, vol.
  2015, 2015.

\bibitem{hauser2013recognition}
K.~Hauser, ``Recognition, prediction, and planning for assisted teleoperation
  of freeform tasks,'' \emph{Autonomous Robots}, vol.~35, no.~4, pp. 241--254,
  2013.

\bibitem{rotter1966generalized}
J.~B. Rotter, ``Generalized expectancies for internal versus external control
  of reinforcement.'' \emph{Psychological monographs: General and applied},
  vol.~80, no.~1, p.~1, 1966.

\bibitem{petrivc2011smooth}
T.~Petri{\v{c}} and L.~{\v{Z}}lajpah, ``Smooth transition between tasks on a
  kinematic control level: Application to self collision avoidance for two kuka
  lwr robots,'' in \emph{2011 IEEE international conference on robotics and
  biomimetics}.\hskip 1em plus 0.5em minus 0.4em\relax IEEE, 2011, pp.
  162--167.

\bibitem{nemec2000null}
B.~Nemec and L.~Zlajpah, ``Null space velocity control with dynamically
  consistent pseudo-inverse,'' \emph{Robotica}, vol.~18, no.~5, pp. 513--518,
  2000.

\bibitem{sadeghian2013task}
H.~Sadeghian, L.~Villani, M.~Keshmiri, and B.~Siciliano, ``Task-space control
  of robot manipulators with null-space compliance,'' \emph{IEEE Transactions
  on Robotics}, vol.~30, no.~2, pp. 493--506, 2013.

\bibitem{henze2016passivity}
B.~Henze, M.~A. Roa, and C.~Ott, ``Passivity-based whole-body balancing for
  torque-controlled humanoid robots in multi-contact scenarios,'' \emph{The
  International Journal of Robotics Research}, vol.~35, no.~12, pp. 1522--1543,
  2016.

\bibitem{abi2018humanoid}
F.~Abi-Farrajl, B.~Henze, A.~Werner, M.~Panzirsch, C.~Ott, and M.~A. Roa,
  ``Humanoid teleoperation using task-relevant haptic feedback,'' in \emph{2018
  IEEE/RSJ International Conference on Intelligent Robots and Systems
  (IROS)}.\hskip 1em plus 0.5em minus 0.4em\relax IEEE, 2018, pp. 5010--5017.

\bibitem{claret2017exploiting}
J.-A. Claret, G.~Venture, and L.~Basa{\~n}ez, ``Exploiting the robot kinematic
  redundancy for emotion conveyance to humans as a lower priority task,''
  \emph{International journal of social robotics}, vol.~9, no.~2, pp. 277--292,
  2017.

\bibitem{Descartes}
\BIBentryALTinterwordspacing
ROS-I. (2015) Descartes—a ros-industrial project for performing path-planning
  on under-defined cartesian trajectories. [Online]. Available:
  \url{http://wiki.ros.org/descartes}
\BIBentrySTDinterwordspacing

\bibitem{malhan2022generation}
R.~K. Malhan, S.~Thakar, A.~M. Kabir, P.~Rajendran, P.~M. Bhatt, and S.~K.
  Gupta, ``Generation of configuration space trajectories over semi-constrained
  cartesian paths for robotic manipulators,'' \emph{IEEE Transactions on
  Automation Science and Engineering}, 2022.

\bibitem{berenson2011task}
D.~Berenson, S.~Srinivasa, and J.~Kuffner, ``Task space regions: A framework
  for pose-constrained manipulation planning,'' \emph{The Int. Journal of
  Robotics Research}, vol.~30, no.~12, pp. 1435--1460, 2011.

\bibitem{cefalo2020opportunistic}
M.~Cefalo, P.~Ferrari, and G.~Oriolo, ``An opportunistic strategy for motion
  planning in the presence of soft task constraints,'' \emph{IEEE Robotics and
  Automation Letters}, vol.~5, no.~4, pp. 6294--6301, 2020.

\bibitem{duttweiler1984internal}
P.~C. Duttweiler, ``The internal control index: A newly developed measure of
  locus of control,'' \emph{Educational and psychological measurement},
  vol.~44, no.~2, pp. 209--221, 1984.

\bibitem{yoshikawa1985manipulability}
T.~Yoshikawa, ``Manipulability of robotic mechanisms,'' \emph{The international
  journal of Robotics Research}, vol.~4, no.~2, pp. 3--9, 1985.

\bibitem{rakita2020effects}
D.~Rakita, B.~Mutlu, and M.~Gleicher, ``Effects of onset latency and robot
  speed delays on mimicry-control teleoperation,'' in \emph{HRI'20: Proceedings
  of the 2020 ACM/IEEE International Conference on Human-Robot Interaction},
  2020.

\bibitem{qiao2021learning}
C.~Z. Qiao, M.~Sakr, K.~Muelling, and H.~Admoni, ``Learning from demonstration
  for real-time user goal prediction and shared assistive control,'' in
  \emph{2021 IEEE International Conference on Robotics and Automation
  (ICRA)}.\hskip 1em plus 0.5em minus 0.4em\relax IEEE, 2021, pp. 3270--3275.

\bibitem{hart1988development}
S.~G. Hart and L.~E. Staveland, ``Development of nasa-tlx (task load index):
  Results of empirical and theoretical research,'' in \emph{Advances in
  psychology}.\hskip 1em plus 0.5em minus 0.4em\relax Elsevier, 1988, vol.~52,
  pp. 139--183.

\bibitem{chiou2021trust}
M.~Chiou, F.~McCabe, M.~Grigoriou, and R.~Stolkin, ``Trust, shared
  understanding and locus of control in mixed-initiative robotic systems,'' in
  \emph{2021 30th IEEE International Conference on Robot \& Human Interactive
  Communication (RO-MAN)}.\hskip 1em plus 0.5em minus 0.4em\relax IEEE, 2021,
  pp. 684--691.

\bibitem{king2018statistical}
B.~M. King, P.~J. Rosopa, and E.~W. Minium, \emph{Statistical reasoning in the
  behavioral sciences}.\hskip 1em plus 0.5em minus 0.4em\relax John Wiley \&
  Sons, 2018.

\bibitem{dragan2013legibility}
A.~D. Dragan, K.~C. Lee, and S.~S. Srinivasa, ``Legibility and predictability
  of robot motion,'' in \emph{2013 8th ACM/IEEE Int. Conf. on Human-Robot
  Interaction (HRI)}.\hskip 1em plus 0.5em minus 0.4em\relax IEEE, 2013, pp.
  301--308.

\bibitem{venture2019robot}
G.~Venture and D.~Kuli{\'c}, ``Robot expressive motions: a survey of generation
  and evaluation methods,'' \emph{ACM Transactions on Human-Robot Interaction
  (THRI)}, vol.~8, no.~4, pp. 1--17, 2019.

\end{thebibliography}
\bibliographystyle{IEEEtran}
\end{document}